\documentclass[letterpaper, 10 pt, conference]{ieeeconf}  % Comment this line out if you need a4paper

\IEEEoverridecommandlockouts                              % This command is only needed if 
\usepackage{graphics} % for pdf, bitmapped graphics files
\usepackage{epsfig} % for postscript graphics files
\usepackage{mathptmx} % assumes new font selection scheme installed
\usepackage{times} % assumes new font selection scheme installed
\usepackage{amsmath} % assumes amsmath package installed
\usepackage{amssymb}  % assumes amsmath package installed
\usepackage{caption}

\usepackage{amsfonts}
\usepackage{multirow}
\usepackage{booktabs}
\usepackage{makecell}
\usepackage{cite}
\usepackage{makecell} % 加入导言区（如未加载）

\usepackage[table]{xcolor} % 必须引入此包以支持 cellcolor
\usepackage{cite}
\usepackage{amsmath,amssymb,amsfonts}
\usepackage{algorithmic}
\usepackage{graphicx}
\usepackage{textcomp}
\usepackage{wrapfig}
\usepackage{epsfig}
\usepackage{mathptmx}
\usepackage{times}
\usepackage{caption}
\usepackage{multirow}
\usepackage{booktabs}
\usepackage{makecell}
\usepackage{cite}
\usepackage{makecell} % 加入导言区（如未加载）
\usepackage[table]{xcolor} % 必须引入此包以支持 cellcolor

\usepackage{booktabs}        % 专业表格线
\usepackage{tabularx}        % 自适应表格宽度
\usepackage{colortbl}        % 表格单元格颜色
\usepackage{adjustbox}       % 表格缩放
\usepackage{array}           % 数组增强

\title{\LARGE \bf
TinyIO: Lightweight Reparameterized Inertial Odometry
}

\author{
Shanshan Zhang$^{1}$, Siyue Wang$^{1}$, Mengzi Chen$^{1}$, Mengzhe Wang$^{1}$,\\ Liqin Wu$^{1}$, Qi Zhang$^{1}$, Lingxiang Zheng$^{1}$
\thanks{$^{1}$Department of Information and Communication Engineering, National and Local Joint Engineering Research Center of Navigation and Location-Based Services, Xiamen University, Xiamen 361005, China.}%
\thanks{*Corresponding authors: Lingxiang Zheng (e-mail: lxzheng@xmu.edu.cn).}%
}

\begin{document}

\maketitle
\thispagestyle{empty}
\pagestyle{empty}

%%%%%%%%%%%%%%%%%%%%%%%%%%%%%%%%%%%%%%%%%%%%%%%%%%%%%%%%%%%%%%%%%%%%%%%%%%%%%%%%
\begin{abstract}

Inertial odometry (IO) is a widely used approach for localization on mobile devices; however, obtaining a lightweight IO model that also achieves high accuracy remains challenging. To address this issue, we propose TinyIO, a lightweight IO method. During training, we adopt a multi-branch architecture to extract diverse motion features more effectively. At inference time, the trained multi-branch model is converted into an equivalent single-path architecture to reduce computational complexity. We further propose a Dual-Path Adaptive Attention mechanism (DPAA), which enhances TinyIO's perception of contextual motion along both channel and temporal dimensions with negligible additional parameters. Extensive experiments on public datasets demonstrate that our method attains a favorable trade-off between accuracy and model size. On the RoNIN dataset, TinyIO reduces the ATE by 23.53\% compared with R-ResNet and decreases the parameter count by 3.68\%.

\textit{Code repository} will be made available upon completion of the double-blind review process.

\end{abstract}

%%%%%%%%%%%%%%%%%%%%%%%%%%%%%%%%%%%%%%%%%%%%%%%%%%%%%%%%%%%%%%%%%%%%%%%%%%%%%%%%
\section{INTRODUCTION}

Inertial odometry (IO) aims to estimate a platform's position over time from inertial measurement unit (IMU) signals~\cite{SurveyofIndoorInertial,surveyILS}. An IMU typically comprises accelerometers and gyroscopes, which measure linear acceleration and angular velocity, respectively. Owing to their low power consumption, independence from external infrastructure, and inherent data privacy preservation, IO has been widely adopted in assistive localization for individuals with disabilities, indoor positioning, and ambient-assisted living~\cite{NILoc,IMUNet,DeepILS,Deeplite,Tartan-IMU,AirIO}.

IO methods can generally be categorized into two types: Newtonian-mechanics-based approaches and data-driven approaches. Traditional methods grounded in Newtonian mechanics offer the advantage of low computational overhead; however, they suffer from cumulative errors and are often constrained by environment-specific factors, which limit their practical applicability~\cite{SINS,PDRusingfrequencydomain,AdaptiveThreshold-BasedZUPT}.

In contrast, data-driven IO methods learn motion patterns directly from IMU signals to estimate position. These methods typically achieve higher localization accuracy and exhibit improved generalization across diverse environments~\cite{Tartan-IMU,AirIO,RIO,WDSNet,EqNIO}. Nevertheless, many existing approaches prioritize accuracy by employing large-scale models while neglecting the need for lightweight architectures suitable for mobile deployment~\cite{IMUNet,DeepILS}.

\begin{figure}[t]
\centering
\includegraphics[width=0.5\textwidth]{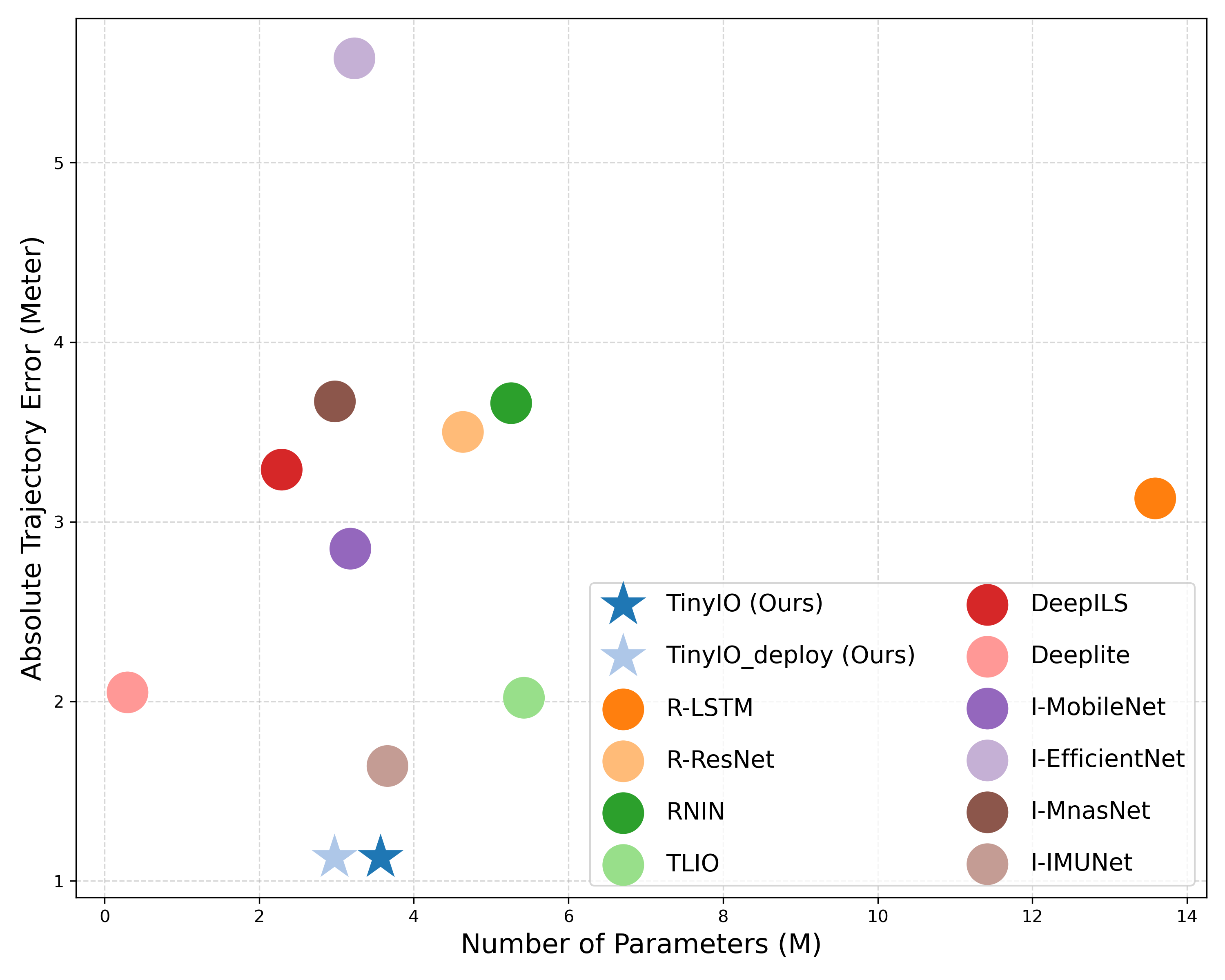}
\caption{Comparison of ATE performance across different models on the OxIOD dataset. Each point corresponds to an IO algorithm; points closer to the bottom-left corner indicate lower trajectory error with fewer model parameters.}
\label{ATE_vs_Params_RNNIN}
\vspace{-18pt}
\end{figure}

To address this issue, we propose a lightweight IO framework named TinyIO, which decouples the training and inference phases to enhance parameter efficiency. The main contributions of this paper are as follows:
\begin{itemize}
    \item We perform inference-time reparameterization to decouple the model from its training architecture, thereby improving parameter efficiency.
    \item We propose a Dual-Path Adaptive Attention mechanism (DPAA) that enhances the network's modeling capability of contextual motion information along both channel and temporal dimensions, introducing only 0.037\,M additional parameters.
    \item Experimental results on multiple public datasets demonstrate that TinyIO achieves a favorable trade-off between accuracy and model size, as illustrated in Fig.~\ref{ATE_vs_Params_RNNIN}.
\end{itemize}

\begin{figure*}[!t]
\centering
\includegraphics[width=1\textwidth]{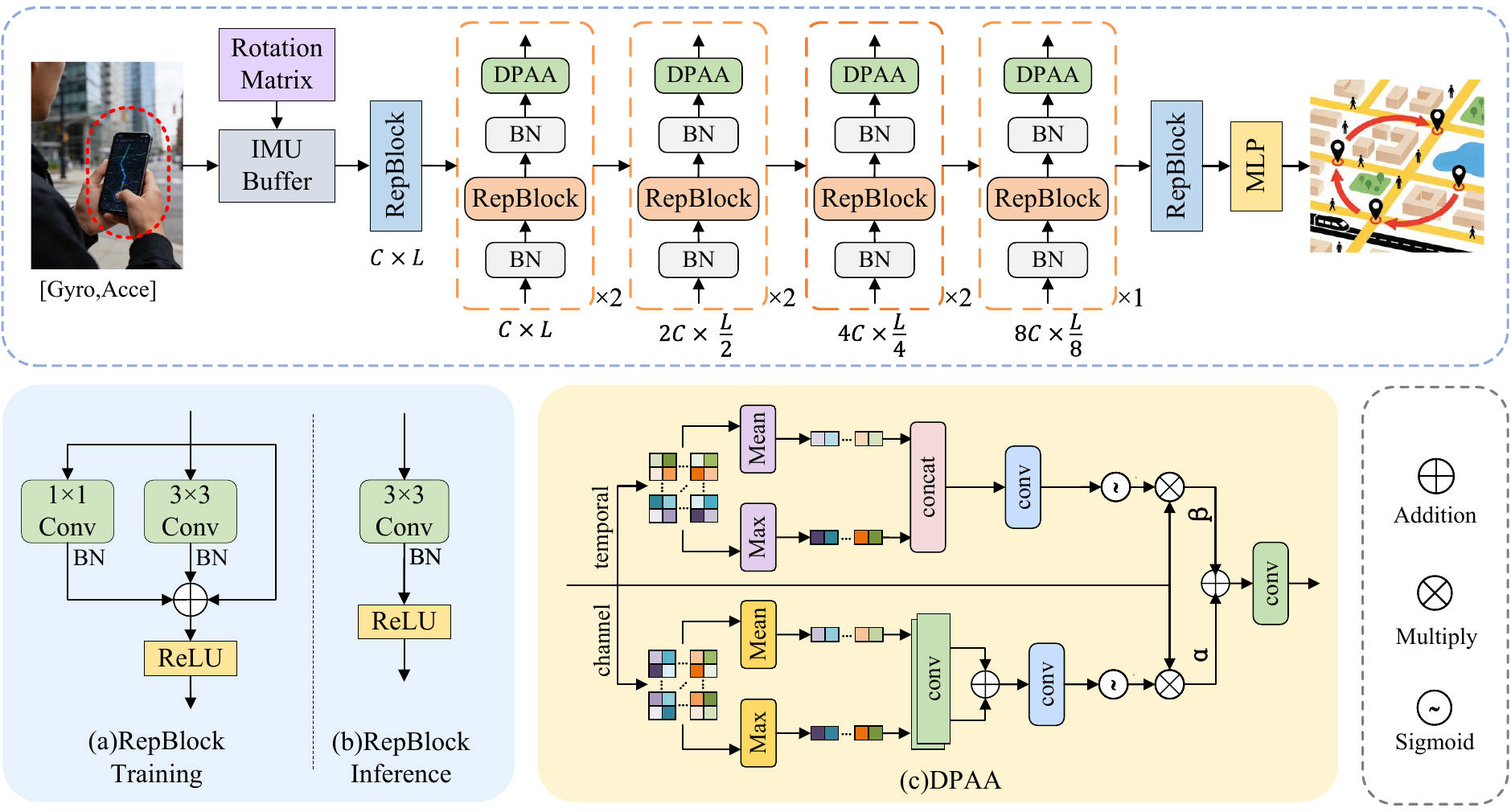}
\caption{The overall architecture of the proposed TinyIO, comprising the Reparameterization Block (RepBlock) and the Dual-path Adaptive Attention mechanism (DPAA).}
\label{framework}
\vspace{-18pt}
\end{figure*}

\section{RELATED WORK}
\subsection{Newtonian-mechanics-based approaches}
Traditional inertial localization methods are typically based on Newtonian mechanics. For example, SINS~\cite{SINS} rotates the acceleration vector into a global coordinate system to compensate for gravity and then estimates displacement through double integration. However, measurement noise continuously accumulates during this process, leading to progressive drift in the estimated trajectory. To mitigate this issue, researchers have introduced physical priors to correct errors. For instance, PDR~\cite{PDR} exploits the periodicity of human motion to detect gait and iteratively updates the trajectory using estimated step length and heading. ZUPT~\cite{ZUPT}, utilizing foot-mounted sensors, applies a zero-velocity constraint during stance phases to correct accumulated errors and improve accuracy under stationary conditions. Although these methods effectively suppress error propagation, their reliance on specific physical priors limits their applicability to a narrow range of motion states and scenarios, reducing robustness in diverse environments.

\subsection{Data-driven approaches}
Data-driven methods have demonstrated superior adaptability across diverse application scenarios, exhibiting greater robustness and improved accuracy compared to traditional approaches\cite{IDOL,RIDI}. For example, IONet~\cite{Ionet} and RoNIN~\cite{RoNIN} disregard the IMU's mounting position and directly infer motion from raw measurements using a unified network framework, irrespective of how the device is worn. Specifically, RoNIN employs ResNet as the inference backbone; its robust localization performance and straightforward architecture have laid a solid foundation for subsequent research. Building on this, TLIO~\cite{TLIO} and LIDR~\cite{LIDR} integrate the SCEKF and the Invariant Extended Kalman Filter (LIEKF), respectively, to further refine ResNet's outputs and mitigate localization errors. RNIN-VIO~\cite{RNIN-VIO} enhances the modeling of long-term dependencies by combining LSTM with R-ResNet. IMUNet\cite{IMUNet}, DeepILS\cite{DeepILS}, and Deeplite\cite{Deeplite} attempt to adopt lightweight architectures to improve parameter efficiency but achieve limited localization accuracy.

In contrast, DIO contends that pose variations resulting from device placement are significant and introduces a novel pose-invariant inertial tracking loss, which regresses both the velocity magnitude and the trigonometric representation of the forward direction~\cite{DIO}. SCHNN~\cite{SCHNN} and SSHNN~\cite{SSHNN} extend DIO by incorporating CNN and LSTM architectures augmented with CNN-based attention mechanisms. Similarly, RIO~\cite{RIO} and EqNIO~\cite{EqNIO} enrich the inference network with components designed to exploit rotational equivariance. More recently, CTIN~\cite{CTIN} and iMOT~\cite{iMOT} have explored the use of Transformer architectures for inertial localization, aiming to capture contextual dependencies, and have achieved promising results.

While the adoption of increasingly complex network architectures has enhanced accuracy, it often comes at the cost of parameter efficiency, posing significant challenges for deployment on resource-constrained devices. To address this, we propose a lightweight localization algorithm that achieves a balance between accuracy and parameter efficiency by incorporating a lightweight attention mechanism and decoupling the training and inference phases.

\section{METHOD}
\subsection{Overall Pipeline}
The overall pipeline of the proposed TinyIO is illustrated in Fig.~\ref{framework}. The TinyIO backbone primarily consists of Lightweight Inertial Odometry Blocks (LIOBs), which infer human motion velocity from acceleration and angular velocity signals. 
% By incorporating a lightweight attention mechanism and decoupling the training and inference phases, the design achieves a balance between accuracy and parameter efficiency.

Each LIOB contains a Reparameterization Blocks (RepBlocks) and a Dual-Path Adaptive Attention mechanism (DPAA). The RepBlock adopts a multi-branch structure during training to capture complex motion patterns. At inference time, this multi-branch structure is reparameterized into an equivalent single-path feedforward form to improve parameter efficiency, as shown in Fig.~\ref{framework}(b). The DPAA compensates for the RepBlock's limited ability to model long-range dependencies by adaptively aggregating motion context along both the temporal and channel dimensions, adhering to the principle of parameter efficiency and introducing only 0.037M additional parameters.

Finally, a multi-layer perceptron (MLP) head predicts velocity, and the network is trained by minimizing the mean squared error (MSE) loss.

\begin{figure}[t]
\centering
\includegraphics[width=0.5\textwidth]{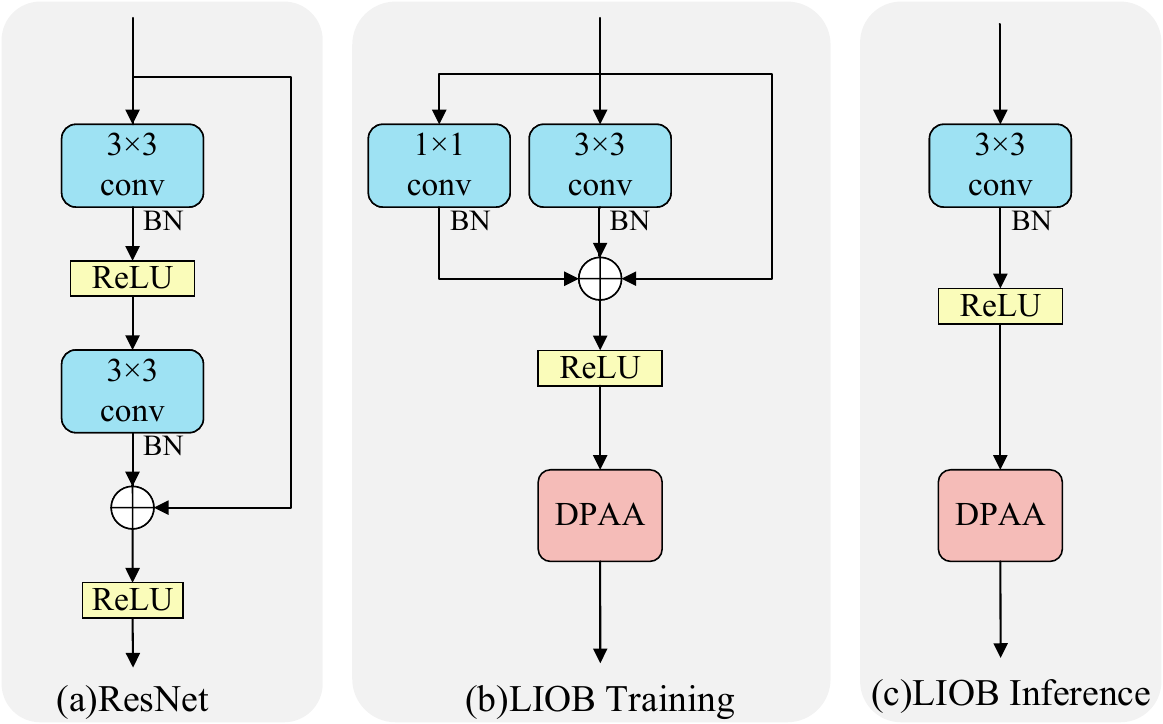}
\caption{Comparison of the basic modules of LIOB and ResNet.}
\label{arc_compare}
\vspace{-18pt}
\end{figure}

\subsection{Lightweight Inertial Odometry Block}
To balance model complexity and accuracy, we incorporate reparameterization techniques into the Lightweight Inertial Odometry Blocks (LIOBs) and propose a dual-path adaptive attention mechanism. Given an IMU signal \( \mathbf{X} \in \mathbb{R}^{C \times F} \), where \( C = 6 \) denotes the six channels from the three-axis accelerometer and gyroscope and \( F \) denotes the number of frames per time window (i.e., the IMU sampling rate), the processing within each LIOB is:
\begin{equation}
\begin{aligned}
\mathbf{X}' &= \mathrm{RepBlock}(\mathbf{X})\\
\mathbf{X}'' &= \mathrm{DPAA}(\mathbf{X}')
\end{aligned}
\end{equation}
Here, \( \mathbf{X}' \) and \( \mathbf{X}'' \) denote the outputs of the RepBlock and the DPAA, respectively.

\textbf{Reparameterization Block (RepBlock).}
The RepBlock is a core component of each LIOB and is also applied at both the beginning and the end of the backbone for channel alignment. To improve parameter efficiency and enhance feature extraction, the RepBlock decouples the training and inference architectures~\cite{RepVGG}. Fig.~\ref{arc_compare} illustrates the structural differences between the RepBlock and the core block used in the R-ResNet\cite{RoNIN} backbone.

During training, a multi-branch architecture—similar to that of ResNet—is adopted to extract features from the IMU signal. Specifically, the computation within each RepBlock during training is given by
\begin{equation}
\mathbf{X}' = \mathbf{X} + \phi(\mathbf{X}) + \psi(\mathbf{X}) \in \mathbb{R}^{C \times F}
\end{equation}
where \( \mathbf{X} \) denotes the input to the block, and \( \phi(\cdot) \) and \( \psi(\cdot) \) denote 1D convolutional branches with kernel sizes 3 and 1, respectively.

At inference, the multi-branch structure is reparameterized into a single-path connection, as shown in Fig.~\ref{arc_compare}(c). This transformation preserves the functional equivalence of the learned representation while simplifying the architecture, thereby improving computational efficiency without degrading predictive accuracy. At inference time, the operation is represented as
\begin{equation}
\mathbf{X}' = \gamma(\mathbf{X}) \in \mathbb{R}^{C \times F}
\end{equation}
where \( \gamma(\cdot) \) denotes a single 1D convolution (kernel size 3) obtained by merging the parameters of \( \phi(\cdot) \), \( \psi(\cdot) \), and the residual (identity) connection from the training phase.

\textbf{Dual-Path Adaptive Attention (DPAA).}
While the RepBlock emphasizes local fine-grained motion features, it lacks sufficient contextual awareness. To address this limitation, we propose the DPAA, which models contextual motion information along both the channel and temporal dimensions through a lightweight attention mechanism and then adaptively fuses them to achieve a comprehensive understanding of the IMU signal. The DPAA structure is illustrated in Fig.~\ref{framework}(d). It comprises two branches: a lightweight channel attention branch that models inter-channel dependencies, and a lightweight temporal attention branch that captures the importance of different time steps.

In the lightweight channel attention branch, we apply adaptive average pooling and adaptive max pooling to the output feature $\mathbf{X}'$ from the RepBlock to extract channel-wise statistics. These statistics are then fused through a two-layer convolutional module with shared parameters and mapped to channel weights in $[0,1]$ via a sigmoid function. The resulting channel weights modulate $\mathbf{X}'$ to emphasize informative channels. Formally, this process is expressed as:
\begin{align}
\mathbf{z}_{\text{avg}} &= \frac{1}{F} \sum_{f=1}^{F} \mathbf{X}'_{:, f} \in \mathbb{R}^{C} \\
\mathbf{z}_{\text{max}} &= \max_{f \in \{1,\dots,F\}} \mathbf{X}'_{:, f} \in \mathbb{R}^{C} \\
\mathbf{w}_c &= \sigma \Big( 
    \mathbf{W}_2 \, \delta ( \mathbf{W}_1 \mathbf{z}_{\text{avg}} ) + 
    \mathbf{W}_2 \, \delta ( \mathbf{W}_1 \mathbf{z}_{\text{max}} )
\Big) \in \mathbb{R}^{C} \\
\mathbf{Y}_{\text{ca}} &= \mathbf{X}' \odot \mathbf{w}_c^{\top} \in \mathbb{R}^{C \times F}
\end{align}
where $\mathbf{z}_{\text{avg}}$ and $\mathbf{z}_{\text{max}}$ denote the channel statistics obtained via adaptive average and max pooling, respectively; $\mathbf{W}_1$ and $\mathbf{W}_2$ are learnable 1D convolutional weights (effectively functioning as shared fully connected layers); $\delta$ and $\sigma$ represent the ReLU and sigmoid activation functions, respectively; $\mathbf{w}_c$ is the fused channel weight vector; and $\odot$ denotes broadcasting element-wise multiplication. Notably, the use of shared-weight convolutions instead of standard fully connected layers further reduces parameter count, satisfying the constraints of resource-constrained deployment.

\begin{table*}[!ht]
\centering
\scriptsize
\setlength{\extrarowheight}{1pt} 
\renewcommand{\arraystretch}{1.0}
\caption{Comparison of trajectory prediction performance and model efficiency. 
ATE and RTE are measured in meters. 
The best results are highlighted with an \textcolor{orange}{orange} background, 
and the second-best results with a \textcolor{green}{green} background.}
\resizebox{\textwidth}{!}{%
% 调整列对齐方式：将所有数值列改为居中 c，保留表头列 l/c
\begin{tabular}{l c c c c c c c c c c c c c |c}
\specialrule{\heavyrulewidth}{0pt}{0pt} 
% ========== 表头第一行：模型名称 ==========
\multirow{2}{*}{\textbf{Dataset}} & 
\multirow{2}{*}{\textbf{Metric}} & 
\multirow{2}{*}{\textbf{R-ResNet}} & 
\multirow{2}{*}{\textbf{R-LSTM}} & 
\multirow{2}{*}{\textbf{RNIN}} & 
\multirow{2}{*}{\textbf{I-IMUNet}} & 
\multirow{2}{*}{\textbf{I-EfficientNet}} & 
\multirow{2}{*}{\textbf{I-MnasNet}} & 
\multirow{2}{*}{\textbf{I-MobileNet}} & 
\multirow{2}{*}{\textbf{DeepILS}} & 
\multirow{2}{*}{\textbf{Deeplite}} & 
\multirow{2}{*}{\textbf{TLIO}} & 
\multirow{2}{*}{\makecell[c]{\textbf{TinyIO\_} \\ \textbf{deploy}}} & 
\multirow{2}{*}{\textbf{TinyIO}} & 
\multirow{2}{*}{\makecell[c]{\textbf{TinyIO vs} \\ \textbf{R-ResNet}}} \\
& & & & & & & & & & & & & & \\
\hline

% ========== 表头第二行：发表信息 ==========
\multicolumn{2}{c}{\textbf{Publication}} 
& \makecell{ICRA \\ 2020}
& \makecell{ICRA \\ 2020}
& \makecell{ISMAR \\ 2021}
& \makecell{TIM \\ 2024} 
& \makecell{TIM \\ 2024} 
& \makecell{TIM \\ 2024} 
& \makecell{TIM \\ 2024} 
& \makecell{IoT-J \\ 2025} 
& \makecell{IoT-J \\ 2025} 
& \makecell{RAL \\ 2020} 
& \makecell{-- \\ --} 
& \makecell{-- \\ --} 
& \makecell{-- \\ --} \\
\hline

% ========== RoNIN 数据集 ==========
\multirow{2}{*}{RoNIN} 
& ATE & \cellcolor{gray!10}6.12 & \cellcolor{gray!10}10.38 & \cellcolor{gray!10}6.88 & \cellcolor{gray!10}6.07 & \cellcolor{gray!10}5.84 & \cellcolor{gray!10}7.49 & \cellcolor{gray!10}4.94 & \cellcolor{gray!10}5.80 & \cellcolor{gray!10}6.36 & \cellcolor{gray!10}5.52 & \cellcolor{orange!25}\textbf{4.55} & \cellcolor{green!15}\textbf{4.68} & \cellcolor{gray!10}23.53\% \\
& RTE & \cellcolor{gray!10}4.08 & \cellcolor{gray!10}4.92 & \cellcolor{gray!10}4.17 & \cellcolor{gray!10}4.04 
& \cellcolor{gray!10}3.91 & \cellcolor{gray!10}4.02 & \cellcolor{green!15}\textbf{3.87} & \cellcolor{gray!10}3.90 & \cellcolor{gray!10}4.38 & \cellcolor{orange!25}\textbf{3.80} & \cellcolor{green!15}\textbf{3.87} & \cellcolor{gray!10}3.93 & \cellcolor{gray!10}3.68\% \\

% ========== RIDI 数据集 ==========
\multirow{2}{*}{RIDI} 
& ATE & \cellcolor{gray!10}2.43 & \cellcolor{gray!10}2.82 & \cellcolor{gray!10}2.77 & \cellcolor{gray!10}3.20 & \cellcolor{green!15}\textbf{1.93} & \cellcolor{gray!10}3.14 & \cellcolor{gray!10}2.89 & \cellcolor{orange!25}\textbf{1.89} & \cellcolor{gray!10}2.31 & \cellcolor{gray!10}2.24 & \cellcolor{orange!25}\textbf{1.89} & \cellcolor{orange!25}\textbf{1.89} & \cellcolor{gray!10}22.22\% \\
& RTE & \cellcolor{gray!10}2.60 & \cellcolor{gray!10}2.98 & \cellcolor{gray!10}2.99 & \cellcolor{gray!10}3.24 & \cellcolor{green!15}\textbf{2.20} & \cellcolor{gray!10}3.00 & \cellcolor{gray!10}3.00 & \cellcolor{gray!10}2.36 & \cellcolor{gray!10}2.56 & \cellcolor{gray!10}2.52 & \cellcolor{orange!25}\textbf{2.19} & \cellcolor{orange!25}\textbf{2.19} & \cellcolor{gray!10}15.77\% \\

% ========== OxIOD 数据集 ==========
\multirow{2}{*}{OxIOD} 
& ATE & \cellcolor{gray!10}3.50 & \cellcolor{gray!10}3.13 & \cellcolor{gray!10}3.66 & \cellcolor{gray!10}1.64 & \cellcolor{gray!10}5.58 & \cellcolor{gray!10}3.67 & \cellcolor{gray!10}2.85 & \cellcolor{gray!10}3.29 & \cellcolor{gray!10}2.05 & \cellcolor{green!15}\textbf{2.02} & \cellcolor{orange!25}\textbf{1.41} & \cellcolor{orange!25}\textbf{1.41} & \cellcolor{gray!10}59.71\% \\
& RTE & \cellcolor{gray!10}1.30 & \cellcolor{gray!10}1.39 & \cellcolor{gray!10}1.34 & \cellcolor{gray!10}1.04 & \cellcolor{gray!10}1.68 & \cellcolor{gray!10}1.46 & \cellcolor{gray!10}1.20 & \cellcolor{gray!10}1.23 & \cellcolor{gray!10}1.11 & \cellcolor{green!15}\textbf{0.97} & \cellcolor{orange!25}\textbf{0.93} & \cellcolor{orange!25}\textbf{0.93} & \cellcolor{gray!10}28.46\% \\

% ========== IMUNet 数据集 ==========
\multirow{2}{*}{IMUNet} 
& ATE & \cellcolor{gray!10}8.16 & \cellcolor{gray!10}12.89 & \cellcolor{gray!10}8.40 & \cellcolor{gray!10}8.68
& \cellcolor{green!15}\textbf{6.11} & \cellcolor{gray!10}12.23 & \cellcolor{gray!10}8.130 & \cellcolor{gray!10}7.86 & \cellcolor{gray!10}7.46 & \cellcolor{gray!10}7.49 & \cellcolor{gray!10}6.35 & \cellcolor{orange!25}\textbf{5.66} & \cellcolor{gray!10}30.64\% \\
& RTE & \cellcolor{gray!10}4.71 & \cellcolor{gray!10}8.71 & \cellcolor{gray!10}5.05 & \cellcolor{gray!10}5.95
& \cellcolor{gray!10}4.60 & \cellcolor{gray!10}7.18 & \cellcolor{gray!10}5.52 & \cellcolor{gray!10}5.59 & \cellcolor{gray!10}5.75 & \cellcolor{gray!10}5.27 & \cellcolor{green!15}\textbf{4.44} & \cellcolor{orange!25}\textbf{4.35} & \cellcolor{gray!10}7.64\% \\
\hline

% 效率指标行
\multicolumn{2}{l}{\textbf{FLOPs (M)}} 
& \cellcolor{gray!10}38.252 & \cellcolor{gray!10}57.473 & \cellcolor{gray!10}72.032 & \cellcolor{green!15}\textbf{18.838} 
& \cellcolor{gray!10}28.831 & \cellcolor{gray!10}25.256
 & \cellcolor{gray!10}34.559 & \cellcolor{orange!25}\textbf{15.287} & \cellcolor{gray!10}59.045 & \cellcolor{gray!10}39.436 & \cellcolor{gray!10}37.595 & \cellcolor{gray!10}26.194 & \cellcolor{gray!10}31.52\% \\
\multicolumn{2}{l}{\textbf{Params (M)}} 
& \cellcolor{gray!10}4.635 & \cellcolor{gray!10}13.589 & \cellcolor{gray!10}5.259 & \cellcolor{gray!10}3.659 
& \cellcolor{gray!10}3.232 & \cellcolor{gray!10}\textbf{2.979} & \cellcolor{gray!10}3.179 & \cellcolor{green!15}2.291 & \cellcolor{orange!25}\textbf{0.297} & \cellcolor{gray!10}5.424 & \cellcolor{gray!10}2.975 & \cellcolor{gray!10}3.570 & \cellcolor{gray!10}22.98\% \\
\multicolumn{2}{l}{\textbf{Peak Memory (MB)}} 
& \cellcolor{gray!10}\textbf{29.069} & \cellcolor{gray!10}80.184 & \cellcolor{gray!10}59.159 & \cellcolor{gray!10}64.955 
& \cellcolor{gray!10}61.256 & \cellcolor{gray!10}72.833 & \cellcolor{gray!10}68.814 & \cellcolor{gray!10}70.979 & \cellcolor{gray!10}50.526 & \cellcolor{gray!10}103.165 & \cellcolor{orange!25}\textbf{22.679} & \cellcolor{green!15}\textbf{25.609} & \cellcolor{gray!10}11.90\% \\
\specialrule{\heavyrulewidth}{0pt}{0pt} 
\end{tabular}%
}
\label{combined_results}
\vspace{-10pt}
\end{table*}

\begin{figure*}[t]
\centering
\includegraphics[width=1\textwidth]{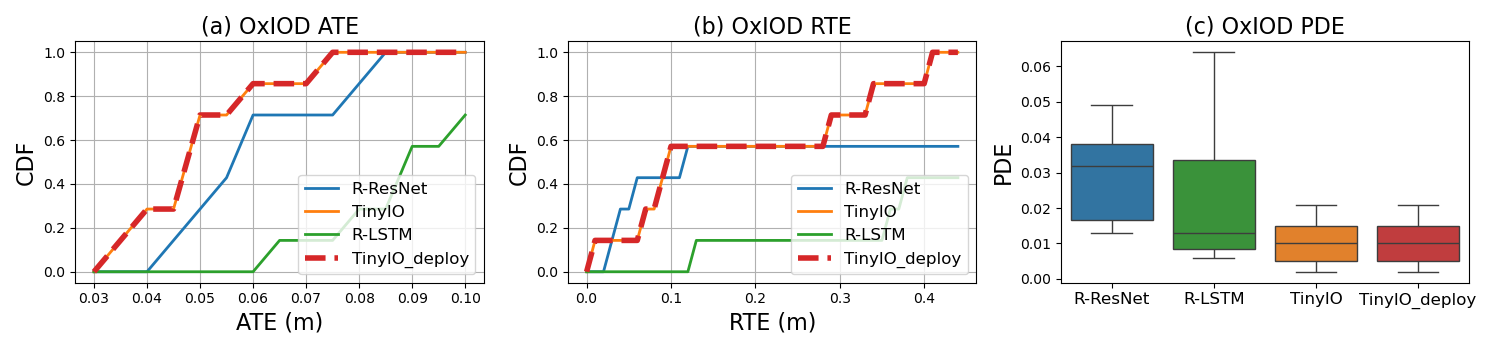}
\caption{(a) and (b) show the cumulative distribution function (CDF) curves of the ATE and RTE for four localization algorithms, respectively. Since TinyIO and TinyIO\_deploy exhibit similar performance, their curves overlap (red and orange curves). (c) presents the PDE metric for the four algorithms.}
\label{PDE_CDF}
\vspace{-15pt}
\end{figure*}

The lightweight temporal attention branch follows a similar design philosophy. We first compute the mean and maximum statistics along the temporal dimension of $\mathbf{X}'$, concatenate them into a two-channel temporal feature map, and then apply a 1D convolution (operating along the time axis) to model local temporal dependencies and map the concatenated features to a temporal attention score. This score is normalized via sigmoid to produce temporal weights in $[0,1]$, which modulate $\mathbf{X}'$ to highlight critical time steps. The procedure is formulated as:
\begin{align}
\mathbf{m}_{\text{avg}} &= \frac{1}{C} \sum_{c=1}^{C} \mathbf{X}'_{c, :} \\
\mathbf{m}_{\text{max}} &= \max_{c \in \{1,\dots,C\}} \mathbf{X}'_{c, :} \\
\mathbf{M} &= 
\begin{bmatrix}
\mathbf{m}_{\text{avg}} \\
\mathbf{m}_{\text{max}}
\end{bmatrix} \\
\mathbf{w}_t &= \sigma \big( \mathbf{W}_3 \mathbf{M} \big) \\
\mathbf{Y}_{\text{ta}} &= \mathbf{X}' \odot \mathbf{w}_t
\end{align}
where $\mathbf{m}_{\text{avg}}$ and $\mathbf{m}_{\text{max}}$ are the temporal mean and max statistics, $\mathbf{W}_3$ is a learnable 1D convolutional kernel applied along the temporal dimension, and $\mathbf{w}_t$ is the resulting temporal weight vector.

Finally, we dynamically fuse $\mathbf{Y}_{\text{ca}}$ and $\mathbf{Y}_{\text{ta}}$ using learnable scalar weights:
\begin{equation}
\begin{aligned}
\mathbf{X}'' &= \alpha \cdot \mathbf{Y}_{\text{ca}} + \beta \cdot \mathbf{Y}_{\text{ta}} \\
1 &= \alpha + \beta
\end{aligned}
\end{equation}
where $\alpha$ and $\beta$ are learnable parameters. This design enables adaptive, dual-path learning over both channel and temporal dimensions while maintaining high parameter efficiency, making it significantly more suitable for mobile deployment compared to standard attention mechanisms.

\section{EXPERIMENTS AND ANALYSIS}
\subsection{Experimental Settings}
\textbf{Training details.}
All training and evaluation are conducted on an NVIDIA RTX 3090 GPU with 24\,GB of memory, using CUDA version 12.2 and PyTorch version 2.5.1. The network is optimized using the Adam optimizer with an initial learning rate of $10^{-4}$. Training runs for up to 40 epochs, with early stopping triggered when the learning rate decays below $10^{-6}$. The channel dimensions of TinyIO's four stages are set to $[48, 96, 192, 512]$.

\textbf{Datasets and metrics.}
We evaluate our method on widely used open-source datasets: RIDI~\cite{RIDI}, RoNIN~\cite{RoNIN}, RNIN~\cite{RNIN-VIO}, and TLIO~\cite{TLIO}. Each dataset is split into training, validation, and test subsets with a ratio of 8:1:1, and all models are retrained on their respective datasets. 
% Furthermore, we deploy the trained model on mobile devices and conduct real-world scenario testing.
The evaluation metrics include commonly used IO metrics: absolute trajectory error (ATE)~\cite{IDOL}, relative trajectory error (RTE)~\cite{IDOL}, and position drift error (PDE)~\cite{CTIN}.

\textbf{Baselines.}
Prior work has demonstrated that data-driven approaches generally outperform Newtonian model-based methods in terms of localization accuracy~\cite{RIDI,Oxiod,RNIN-VIO,TLIO,iMOT}. Accordingly, we compare TinyIO against widely used open-source data-driven baselines, including DeepILS~\cite{DeepILS}, Deeplite~\cite{Deeplite}, the R-ResNet and R-LSTM variants from RoNIN~\cite{RoNIN}, the I-IMUNet, I-MobileNet, I-MNasNet, and I-EfficientNet models from IMUNet~\cite{IMUNet}, as well as the architectures proposed in TLIO~\cite{TLIO} and RNIN~\cite{RNIN-VIO}.

\begin{figure*}[!t]
\centering
\includegraphics[width=1\textwidth]{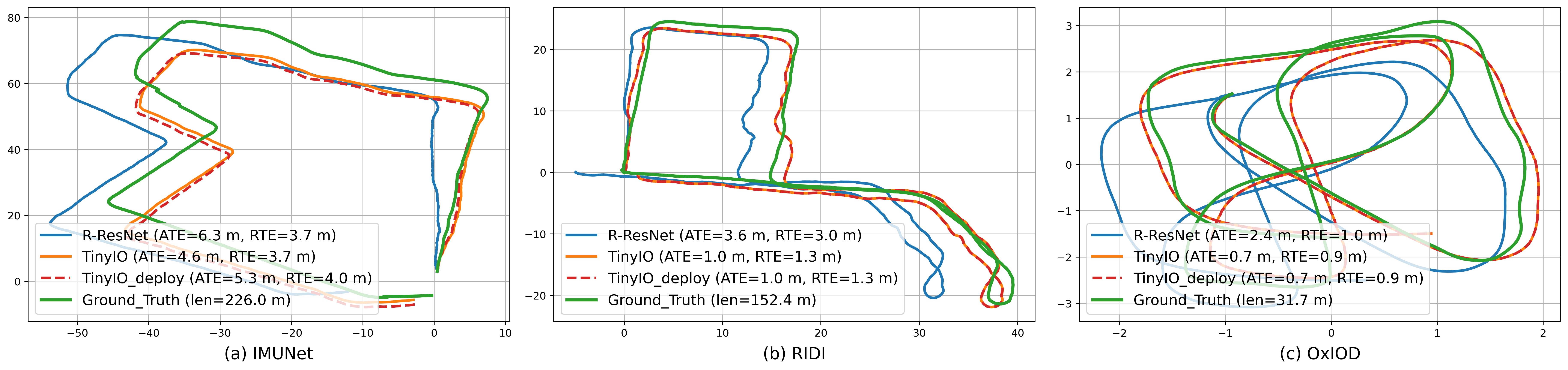}
\caption{
Comparison of sampled trajectories on the test set for TinyIO, TinyIO\_deploy, and R-ResNet. The values in parentheses denote the specific metrics for each algorithm on the corresponding trajectory. Due to the similar performance of TinyIO and TinyIO\_deploy, their trajectories exhibit overlapping coverage (red and orange curves). For the OxIOD dataset, only a segment of the full trajectory is visualized to fit the display scale.
}
\label{combined_plot}
\vspace{-10pt}
\end{figure*}

\subsection{Comparisons with the baselines}
\textbf{Metric evaluation.}
Table~\ref{combined_results} summarizes ATE and RTE comparisons between TinyIO and the baseline models. TinyIO and its reparameterized variant, TinyIO\_deploy, achieve the lowest localization errors in most scenarios. For example, relative to R-ResNet, TinyIO reduces ATE by 22.22\%\textemdash59.71\% and RTE by 3.68\%\textemdash28.46\% across the all datasets. Model complexity increases progressively from RoNIN to TLIO and RNIN, which raises deployment challenges for edge devices. Although IMUNet uses depthwise-separable convolutions to reduce complexity, its localization accuracy remains limited. In contrast, TinyIO decouples training and inference to substantially reduce parameter counts, while the DPAA models contextual motion information through a lightweight attention mechanism to strike an effective balance between accuracy and compactness.
Overall, these improvements explain the superior performance of TinyIO over the baselines.

\textbf{Visualization.}
Fig.~\ref{PDE_CDF}(a) and (b) present the cumulative distribution function (CDF) curves of the ATE and RTE for  TinyIO, TinyIO\_deploy, R-LSTM and R-ResNet on the IMUNet dataset, respectively. Results on the IMUNet dataset are presented due to space constraints. As shown in the figures, the TinyIO curve (orange) is positioned above and to the left of the R-ResNet curve (blue) and R-LSTM (green), indicating that TinyIO achieves higher accuracy compared to R-ResNet and R-LSTM. For instance, approximately 82\% of trajectories from TinyIO exhibit ATE values below 0.06\,m, whereas only approximately 72\% of trajectories from R-ResNet achieve ATE values below this threshold.
Fig.~\ref{PDE_CDF}(c) shows a comparison of PDE. TinyIO attains the smallest PDE median and the most compact error distribution, indicating superior endpoint approximation. Although reparameterization slightly increases the PDE for TinyIO\_deploy, it still outperforms R-ResNet and preserves high-fidelity trajectory reconstruction consistent with the training-phase model.

\begin{table}[!tbp]
    \centering
    \renewcommand{\arraystretch}{1.2}
    \caption{Ablation study results on open-source datasets. 
             Best results are highlighted in \textcolor{orange}{orange}, 
             second-best in \textcolor{green}{green}.}
    \label{tab:ablation_study}
    
    \begin{adjustbox}{width=\linewidth,center}
    \begin{tabular}{l l c c c}
        \toprule[1.2pt]
        \multirow{2}{*}{\textbf{Dataset}} 
        & \multirow{2}{*}{\textbf{Metric}} 
        & \textbf{R-ResNet} 
        & \begin{tabular}[c]{@{}c@{}}\textbf{TinyIO} \\ \scriptsize(only RepBlock)\end{tabular} 
        & \begin{tabular}[c]{@{}c@{}}\textbf{Full TinyIO} \\ \scriptsize(RepBlock\&DPAA)\end{tabular} \\
        \cmidrule(lr){3-5}
        & & \multicolumn{3}{c}{Performance (meters)} \\
        \hline
        
        % RoNIN Dataset
        \multirow{2}{*}{RoNIN}   
        & ATE  & \cellcolor{gray!10}6.12  & \cellcolor{green!15}\textbf{4.99}  & \cellcolor{orange!25}\textbf{4.68} \\
        & RTE  & \cellcolor{gray!10}4.08  & \cellcolor{green!15}\textbf{3.94}  & \cellcolor{orange!25}\textbf{3.93} \\
        
        % RIDI Dataset
        \multirow{2}{*}{RIDI}    
        & ATE  & \cellcolor{gray!10}2.43  & \cellcolor{green!15}\textbf{2.16}
  & \cellcolor{orange!25}\textbf{1.89} \\
        & RTE  & \cellcolor{gray!10}2.60  & \cellcolor{green!15}\textbf{2.50}  & \cellcolor{orange!25}\textbf{2.19} \\
        
        % IMUNet Dataset
        \multirow{2}{*}{IMUNet}  
        & ATE  & \cellcolor{gray!10}8.16  & \cellcolor{green!15}\textbf{7.45} & \cellcolor{orange!25}\textbf{5.66} \\
        & RTE  & \cellcolor{gray!10}4.71  & \cellcolor{green!15}\textbf{5.13}  & \cellcolor{orange!25}\textbf{4.35} \\
        
        % OxIOD Dataset
        \multirow{2}{*}{OxIOD}   
        & ATE  & \cellcolor{gray!10}3.50  & \cellcolor{green!15}\textbf{2.28}  & \cellcolor{orange!25}\textbf{1.41} \\
        & RTE  & \cellcolor{gray!10}1.30  & \cellcolor{green!15}\textbf{1.12}  & \cellcolor{orange!25}\textbf{0.93} \\
        \hline
        
        % Model Parameters
        \multicolumn{2}{l}{Parameters (M)}      
        & \cellcolor{gray!10}4.635 & \cellcolor{orange!25}\textbf{3.533} & \cellcolor{green!15}\textbf{3.57}
 \\
        \multicolumn{2}{l}{FLOPs (M)}               
        & \cellcolor{gray!10}38.252 & \cellcolor{orange!25}\textbf{26.019} & \cellcolor{green!15}\textbf{26.194}
 \\
        \multicolumn{2}{l}{Peak Memory (MB)}        
        & \cellcolor{gray!10}29.069
 & \cellcolor{orange!25}\textbf{25.587} & \cellcolor{green!15}\textbf{25.609}
 \\
        \specialrule{\heavyrulewidth}{0pt}{0pt} 
    \end{tabular}
    \end{adjustbox}
\label{ablation}
\vspace{-15pt}
\end{table}

Fig.~\ref{combined_plot} visualizes trajectory reconstructions produced by TinyIO and the baseline models on the test sets. In Fig.~\ref{combined_plot}(a), TinyIO effectively captures contextual motion information, yielding accurate and stable tracking. After reparameterization, the positioning error of TinyIO\_deploy increases slightly but remains lower than that of R-ResNet, suggesting that the accuracy degradation due to reparameterization is within acceptable bounds. For complex trajectories (Fig.~\ref{combined_plot}(b), (c)), TinyIO better models local fine-grained motion and exhibits substantially less drift than R-ResNet when handling rotational or highly nonlinear motions; consequently, both TinyIO and TinyIO\_deploy closely follow ground-truth trajectories, whereas R-ResNet shows pronounced drift during turns and increasingly larger deviations as trajectory complexity grows.

% \subsection{Real-World Scenarios}
% To further evaluate the practical performance of TinyIO, we deployed it on a smartphone and conducted real-world experiments. In our setup, we adopted a data collection protocol similar to that of RoNIN, as illustrated in Figure~5. An iPhone~17 was used as a fixed ground-truth tracking device, with trajectory ground truth recorded using the RTAB-Map algorithm available on the Apple App Store. Meanwhile, TinyIO and the baseline methods were deployed on a randomly held smartphone to simulate realistic usage conditions and perform real-time localization. A video clip of the experimental procedure is provided in the supplementary materials. As shown in Table~2 and Figure~6, among algorithms designed with lightweight objectives, TinyIO achieves the highest localization accuracy. In terms of inference speed, although TinyIO is slightly slower than Deeplite, it remains well within practical requirements, thereby achieving a favorable balance between accuracy and computational efficiency.

\subsection{Ablation study}
We conduct ablation experiments to validate the reparameterization strategy and the contributions of individual modules (DPAA and RepBlock).

The  Table~\ref{combined_results} reports performance comparisons between TinyIO and TinyIO\_deploy across the all datasets. After reparameterization, TinyIO\_deploy (2.975M parameters) reduces the parameter count by approximately 16.67\% relative to TinyIO (3.570M), and its model size is only 64.19\% that of R-ResNet (4.635M). Despite this compression, the increases in ATE and RTE for TinyIO\_deploy are marginal, and TinyIO\_deploy still outperforms R-ResNet in localization accuracy. Visualization results in Fig.~\ref{combined_plot} corroborate these findings. These results indicate that the proposed reparameterization achieves a favorable trade-off between parameters and accuracy, making TinyIO particularly suitable for edge devices with limited storage.

The Table~\ref{ablation} shows module ablation results. Removing DPAA (TinyIO (only RepBlock)) leaves only the RepBlock: the model still captures local fine-grained motion and yields substantially lower localization error than R-ResNet, indicating that RepBlock both reduces parameter complexity and contributes positively to accuracy. 
Integrating DPAA into the full TinyIO (RepBlock\&DPAA) further improves accuracy, indicating that modeling contextual motion through DPAA is essential for enhancing trajectory estimation. These findings confirm the effectiveness and complementarity of the proposed components.

\section{Conclusion}
We present TinyIO, a lightweight IO framework that attains a favorable trade-off between localization accuracy and model parameters. In practical deployment, however, additional procedures such as model fine-tuning may be necessary to adapt to previously unseen carrier motion patterns.

\bibliographystyle{IEEEtran}
\bibliography{ref}
\end{document}